\documentclass[letterpaper]{article} 
\usepackage{aaai2027}  
\nocopyright
\usepackage[hyphens]{url}  
\usepackage{graphicx} 
\usepackage{natbib}  
\usepackage{caption} 
\usepackage{algorithm}
\usepackage{algorithmic}
\usepackage{comment}
\usepackage{amsthm}

\usepackage{amsfonts}

\usepackage{booktabs}
\usepackage{graphicx}
\usepackage[table]{xcolor}
\usepackage{courier}
\usepackage{newfloat}
\usepackage{amsthm}
\usepackage{amsmath}
\usepackage{listings}
\usepackage{cleveref}
\DeclareCaptionStyle{ruled}{labelfont=normalfont,labelsep=colon,strut=off} 
\floatstyle{ruled}
\newfloat{listing}{tb}{lst}{}
\floatname{listing}{Listing}

\usepackage{booktabs}

\title{\texttt{SymKL}: A Unified Representation-Space Measure of Signal Degradation}

\author{
    Dhananjaya Jayasundara\textsuperscript{*},
    Sudarshan Rajagopalan\textsuperscript{*},
    Vishal M. Patel\textsuperscript{*}
}

\affiliations{}

\begin{document}
\maketitle

\begingroup
\renewcommand{\thefootnote}{*}
\footnotetext{Johns Hopkins University. Corresponding author: vjayasu1@jh.edu}
\endgroup

\begin{abstract}
Quantifying degradation between a pristine signal and its degraded observation is fundamental to signal processing, yet existing full-reference measures are inherently modality-specific. We introduce \textbf{\texttt{SymKL}}, a unified full-reference degradation measure based on implicit neural representations (INRs). An INR is fitted once to the pristine reference and characterized by its diagonal Fisher information, which captures the local information geometry of the learned representation. Evaluating the Fisher information with the pristine reference and a degraded observation yields two parameter-sensitivity profiles whose symmetric KL divergence defines the degradation score. The proposed formulation requires only the reference and degraded observation, without knowledge of the corruption process or degradation severity, making it applicable across images, audio, and 3D data. We further derive an exact closed-form expression for the diagonal Fisher of SIREN, enabling efficient computation. Experiments on image, speech, and volumetric 3D benchmarks demonstrate that \textbf{\texttt{SymKL}} accurately tracks degradation severity, remains monotonic where conventional metrics saturate or fail, and provides a unified modality-agnostic measure of signal degradation. Code will be made publicly available after the review process.
\end{abstract}


\section{Introduction}

\begin{figure}[t]
    \centering
    \includegraphics[width=\columnwidth]{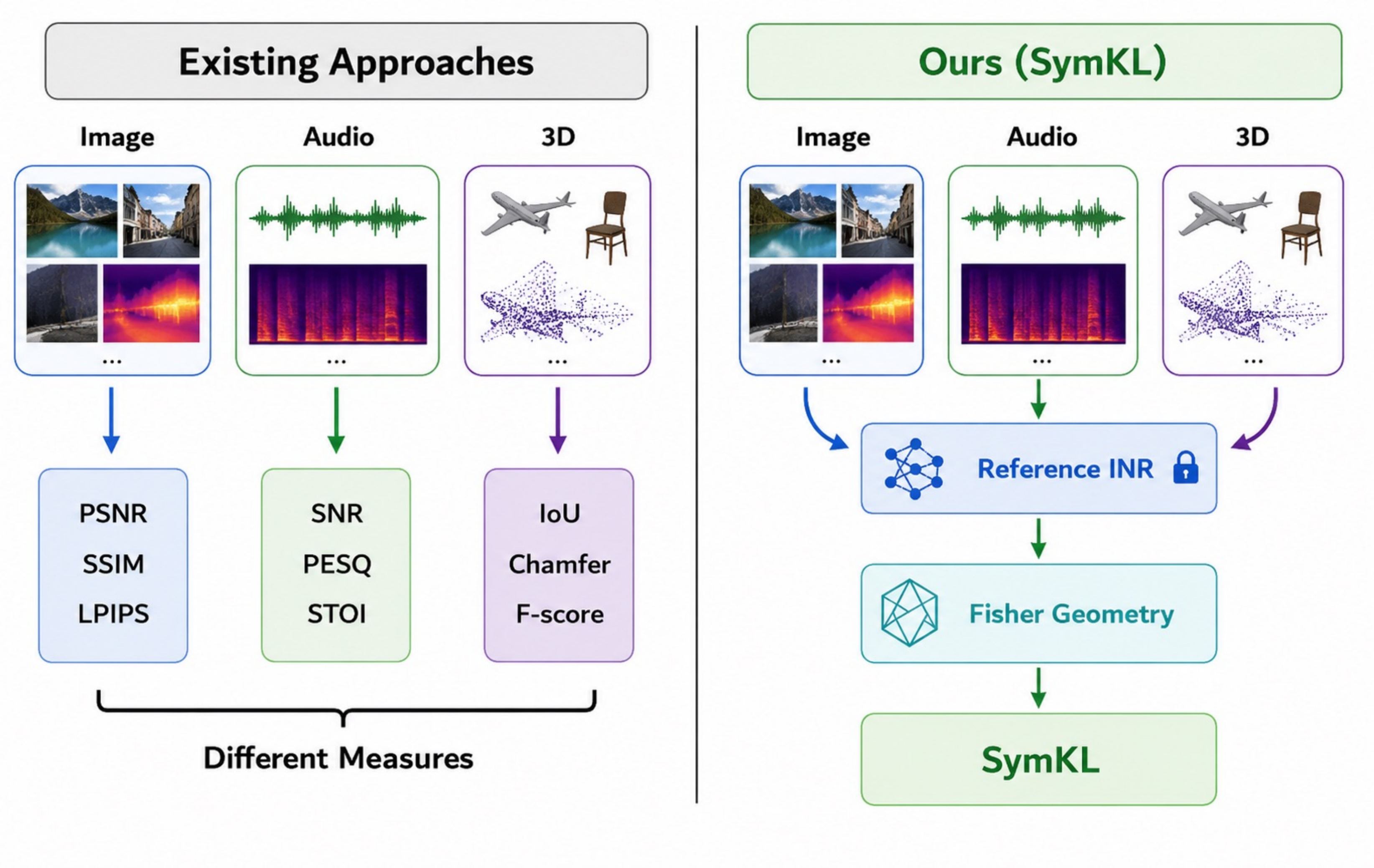}
\caption{\textbf{Conceptual comparison between existing degradation measures and the proposed \textbf{\texttt{SymKL}}.} Existing approaches rely on modality-specific measures, requiring separate metrics for images, audio, and 3D data. In contrast, \textbf{\texttt{SymKL}} represents signals using an INR and characterizes degradation through Fisher information geometry, yielding a unified degradation measure across modalities.}
    \label{fig:concept}
\end{figure}

Signal degradation is an inevitable consequence of acquisition, transmission, compression, and reconstruction processes, making its accurate quantification a fundamental problem in signal processing. Reliable measures of reconstruction fidelity are indispensable for developing restoration algorithms, benchmarking competing methods, and evaluating reconstruction quality across a wide range of applications. Consequently, numerous degradation measures have been developed for different signal modalities. Images are commonly evaluated using pixel-fidelity measures such as PSNR \cite{huynh2008scope}, structural similarity measures including SSIM \cite{nilsson2020understanding} and MS-SSIM \cite{rouse2008understanding}, and learned perceptual measures such as LPIPS \cite{zhang2018unreasonable} and DreamSim \cite{fu2023dreamsim}. Audio quality is typically assessed using signal-fidelity measures including SNR \cite{johnson2006signal} and SI-SDR \cite{le2019sdr}, together with perceptual measures such as PESQ \cite{rix2001perceptual} and STOI \cite{zhang2018training}, while 3D representations are commonly evaluated using geometric measures such as Chamfer Distance \cite{wu2021density}, Intersection-over-Union (IoU) \cite{zhou2019iou}, and the F-score \cite{tatarchenko2019single}. Collectively, these measures have become the standard tools for evaluating reconstruction quality within their respective domains.

Despite their success, these methods have largely evolved independently for different signal modalities and are built upon fundamentally different representations and similarity criteria. Image domain measures compare pixels, structural statistics, or learned visual features, audio measures evaluate waveform fidelity or perceptual intelligibility, while geometric measures assess spatial correspondence between shapes. Consequently, different modalities require fundamentally different degradation algorithms, each tailored to the characteristics of its underlying representation. Moreover, even within a single modality, the response of existing measures may become progressively less informative over parts of the degradation spectrum. For example, on the non-eccentricity distortion of KADID-10k \cite{lin2019kadid}, increasing the number of displaced image patches by a factor of five changes CLIP \cite{radford2021learning} similarity by less than $1\%$ (from $0.998$ to $0.990$), while the visual distortion increases substantially. A similar phenomenon can be observed in 3D geometry, where the mesh F-score collapses to zero once the surface displacement exceeds its matching tolerance, despite geometric differences continuing to increase thereafter (See Supplementary). These observations motivate a fundamental question, \emph{is there a unified representation in which signal degradation can be characterized consistently across different modalities while preserving sensitivity to distortion severity?}

Implicit neural representations (INRs) provide a promising foundation for addressing this question. Rather than representing signals as pixels, waveform samples, or geometric primitives, INRs learn continuous functions from discrete observations using a common coordinate-based neural network formulation. The same representation has been successfully applied to images, audio, videos, occupancy fields, signed-distance fields, neural radiance fields, and many other signal modalities. Unlike conventional degradation measures, which require different algorithms for different data types, INRs provide a unified representation capable of modeling a broad range of signals \cite{essakine2024we, jayasundara2026implicit}. This naturally leads to the central question of this work, \emph{how is signal degradation encoded within this shared neural representation?}

Directly comparing INR parameters is uninformative because different parameters contribute unequally to representing the signal. Instead, \textbf{\texttt{SymKL}} compares their functional importance, quantified by the Fisher information \cite{ly2017tutorial}, which measures the sensitivity of the reconstruction objective to parameter perturbations. A reference INR is fitted once to the pristine signal and then frozen. Evaluating its Fisher information using the pristine and degraded observations yields two parameter-sensitivity profiles whose discrepancy defines a modality-agnostic representation-space measure of degradation (see \Cref{fig:concept}).This naturally raises the question of why we characterize the representation using the Fisher information rather than gradients or the Hessian. The mean gradient vanishes near a fitted solution and therefore provides little information about the learned representation, while the Hessian, although capturing second-order curvature, can be computationally prohibitive for large INRs. The Fisher information retains the required second-order information while for SIREN networks, admits the exact closed-form expression.

Making the \textbf{\texttt{SymKL}} framework practical requires overcoming two computational challenges. Computing Fisher information through conventional automatic differentiation can be prohibitively expensive for large INRs. We therefore derive an exact closed-form expression for the diagonal Fisher information of SIREN networks \cite{sitzmann2020implicit}, reducing the computation to efficient matrix operations while preserving exactness. In addition, fitting a high-quality INR independently for every reference signal is computationally demanding. To address this limitation, we develop an amortized fitting framework based on a hypernetwork \cite{chauhan2024brief} with first-order MAML \cite{nichol2018first} adaptation that rapidly produces accurate reference representations using only a small number of optimization steps. Finally, we introduce a physically grounded evaluation protocol that compares heterogeneous degradation measures according to their responsiveness to continuously varying physical distortion severity. Extensive experiments on natural images, audio, occupancy fields, and medical imaging demonstrate that the \textbf{\texttt{SymKL}} consistently tracks degradation severity, remains discriminative in regimes where conventional degradation measures become less informative, and substantially outperforms alternative weight-space characterizations.


Note that \textbf{\texttt{SymKL}} is not a perceptual quality metric.  Unlike perceptual quality measures, our objective is not to approximate human judgments of visual or auditory quality. Instead, \textbf{\texttt{SymKL}} measures how strongly degradation alters the internal organization of a learned neural representation. Consequently, it is applicable to signal modalities where perceptual judgments are ill-defined, unavailable, or not the primary objective, including volumetric representations, occupancy fields, medical imaging, and scientific measurements.

We make the following contributions.  1) We introduce the first modality-agnostic degradation measure based on the information geometry of implicit neural representations.  2) We show that degradation can be characterized through changes in parameter sensitivities rather than observation-space discrepancies.  3) We derive an exact closed-form Fisher computation for SIREN networks, enabling efficient evaluation. 4) We demonstrate a single degradation measure across images, audio, and 3D signals.

\section{Method}
\label{sec:method}

\begin{figure}[t]
    \centering
    \includegraphics[width=\columnwidth]{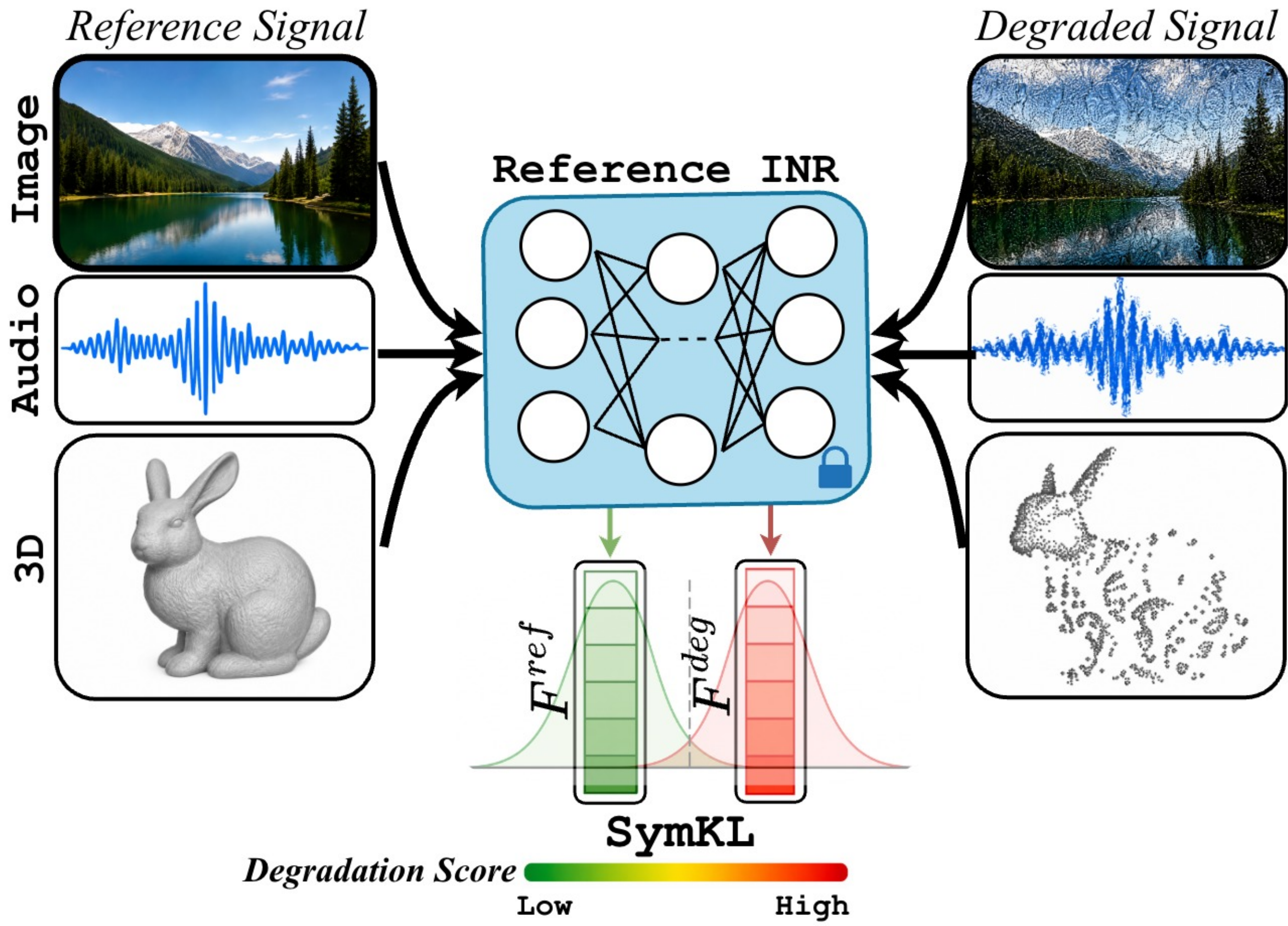}
\caption{\textbf{Overview of \texttt{SymKL}.} Given a reference signal, \textbf{\texttt{SymKL}} first fits an INR to the reference and subsequently freezes its parameters. The frozen reference INR is then evaluated on both the reference and degraded signals to obtain their Fisher information representations. The symmetric KL divergence between these Fisher representations yields a universal degradation score that consistently quantifies degradation across diverse signal modalities}
    \label{fig:main_figure}
\end{figure}

\vspace{-5pt}

\subsection{Preliminaries: signals as coordinate networks}
\label{sec:method:prelim}

Let a signal be a mapping $x:\Omega\to\mathbb{R}^{C}$ on a coordinate domain $\Omega\subset[-1,1]^{D}$, sampled at $N$ coordinates $\{u_n\}_{n=1}^{N}$: an image ($D{=}2$, $C{=}3$), a waveform ($D{=}1$, $C{=}1$), or an occupancy/signed-distance volume ($D{=}3$, $C{=}1$). We represent $x$ using a SIREN, an $L$-layer perceptron $f_\theta:\Omega\to\mathbb{R}^{C}$ with sinusoidal activations, where $h^{(0)}=u$, $z^{(\ell)}=W_\ell h^{(\ell-1)}+b_\ell$, and $h^{(\ell)}=\sin(\omega_\ell z^{(\ell)})$ for $\ell=1,\ldots,L-1$, followed by the linear output layer $f_\theta(u)=W_Lh^{(L-1)}+b_L$. The network parameters are $\theta=\{W_\ell,b_\ell\}_{\ell=1}^{L}\in\mathbb{R}^{P}$. The frequency scales $\omega_\ell$ are modality-dependent \cite{sitzmann2020implicit} ; importantly, the proposed degradation measure is identical across modalities. The INR is fitted by minimizing the mean per-coordinate reconstruction loss $\mathcal{L}(\theta;x)=\frac{1}{N}\sum_{n=1}^{N}\ell_n(\theta;x)$, where $\ell_n(\theta;x)=\frac{1}{2C}\|f_\theta(u_n)-x(u_n)\|_2^2$.

\subsection{Amortized reference fitting for images}
\label{sec:method:hyper}

Obtaining the reference INR through optimization can be computationally expensive \cite{sitzmann2020implicit, saragadam2023wire, ramasinghe2022beyond}. We therefore employ a hypernetwork $\Phi$ to predict the parameters of the target SIREN directly from the reference image, providing a high-quality initialization that substantially reduces the optimization required to obtain the reference representation. The proposed degradation measure is defined solely by the final reference representation and is therefore agnostic to how that representation is obtained. 
For modalities where no pretrained hypernetwork is available, such as audio and volumetric data, the reference INR is obtained through direct optimization.

The hypernetwork consists of three components: a DINOv2 ViT-B/14
encoder~\cite{oquab2024dinov2}, a transformer decoder, and lightweight
prediction heads. Given a reference image, the encoder extracts dense patch
features. The decoder maintains one learned query token per neuron of the
target SIREN, which alternately attend to one another and cross-attend to the
patch features, yielding a latent representation for each neuron. The
prediction heads then map these latents to the corresponding weights and
biases of the SIREN. We observed that directly predicting the network parameters is unstable, with
unconstrained outputs the head activations grow without bound while the output
scale shrinks to compensate, and training diverges (see Supplementary). Instead, the heads estimate
bounded residuals with respect to a learnable base network,
$\hat W_\ell = W_\ell^{\mathrm{base}} + \beta_\ell^{W}\tanh(s\,o_\ell^{W})$ and
$\hat b_\ell = b_\ell^{\mathrm{base}} + \beta_\ell^{b}\tanh(s\,o_\ell^{b})$,
where $o_\ell^{W}, o_\ell^{b}$ are the head outputs,
$W_\ell^{\mathrm{base}}, b_\ell^{\mathrm{base}}$ are learnable base parameters
initialized by the SIREN scheme, and $s$ is a learned scale. The weight bounds
$\beta_\ell^{W}$ are set to SIREN's own initialization bounds, and the bias
bounds $\beta_\ell^{b}$ to one radian of phase, so predictions remain within a
factor of two of a valid initialization. The heads are zero-initialized, so
optimization begins exactly at the base network and learns bounded corrections
to it rather than unconstrained parameters, substantially improving stability.

Training proceeds in two stages. We first pretrain the hypernetwork for image reconstruction on the combined MS-COCO~\cite{awsaf2020coco2017} and LSDIR~\cite{li2023lsdir} datasets, enabling it to predict high-quality initializations for the target SIREN. The network is then fine-tuned using a first-order MAML objective~\cite{finn2017model} so that the predicted parameters can be rapidly adapted to each reference image through a small number of reconstruction steps. During inference, the predicted parameters are refined. Additional architectural details, training schedules, optimization settings, and implementation details are provided in the supplementary material.

\subsection{Fisher profiles of a frozen representation}
\label{sec:method:fisher}

Once the reference INR has been fitted, its parameters are frozen and are never
updated again. Given this fixed representation, our goal is to characterize how
differently it responds to the pristine signal and to its degraded counterpart,
without learning a separate network for the degraded observation. Instead of
comparing reconstructed signals or fitting a second INR, we measure how
sensitive the parameters of the same reference representation are to each
signal, quantified by the diagonal empirical Fisher information evaluated at
the frozen parameters $\theta_{\mathrm{ref}}$ (See \Cref{fig:main_figure}).

For any target signal $y$ (either the pristine reference or a degraded observation defined on the same coordinates), let $g_n(y)=\nabla_\theta \ell_n(\theta;y)\rvert_{\theta=\theta_{\mathrm{ref}}}$ denote the reconstruction gradient at coordinate $u_n$. The gradient used for reconstruction optimization is the mean of these per-coordinate gradients, $\bar g(y)=\frac{1}{N}\sum_{n=1}^{N}g_n(y)$ (as in Preliminaries section). Instead of this first-order quantity, we consider their diagonal second moment,
$F(y)=\frac{1}{N}\sum_{n=1}^{N}g_n(y)\odot g_n(y)$, corresponding to the diagonal empirical Fisher information. Intuitively, $F(y)$ measures the average squared sensitivity of each network parameter to the target signal. Near a fitted solution, the positive and negative contributions to the mean gradient largely cancel, causing $\bar g(y)$ to approach zero even though individual coordinates may still induce substantial parameter sensitivities. In contrast, the Fisher information accumulates squared gradients and therefore preserves the contribution of every coordinate. Empirically, we observe that $\|\sqrt{F}\|$ exceeds $\|\bar g\|$ by more than two orders of magnitude at $\theta_{\mathrm{ref}}$, illustrating the extent of this cancellation. Consequently, $F(y)$ provides a substantially richer per-parameter sensitivity profile for characterizing how the frozen representation explains the target signal. Following standard practice in continual learning~\cite{kirkpatrick2017overcoming} and task embedding~\cite{achille2019task2vec}, we treat this diagonal profile as the precision of a Laplace-type Gaussian over the parameters, $p_y(\theta)=\mathcal{N}\!\left(\theta_{\mathrm{ref}},\,\operatorname{diag}(F(y))^{-1}\right)$. Crucially, all such distributions share the same mean $\theta_{\mathrm{ref}}$, since the reference representation is never updated.

\subsection{The degradation measure}
\label{sec:method:kl}

The previous section associates every target signal with a Fisher profile describing the parameter sensitivities of the frozen reference representation. Given the reference profile $F^{\mathrm{ref}}=F(x_{\mathrm{ref}})$ and the degraded profile $F^{\mathrm{deg}}=F(x_{\mathrm{deg}})$, our goal is to quantify how differently the two signals are represented. As each Fisher profile defines the precision of a Laplace Gaussian, this reduces to comparing two probability distributions over the same parameter space. A natural measure of the discrepancy between two probability distributions is the Kullback--Leibler (KL) divergence \cite{raiber2017kullback}. For diagonal Gaussians with a shared mean, the KL divergence has the closed-form expression $\mathrm{KL}(F^{\mathrm{ref}}\Vert F^{\mathrm{deg}})=\frac{1}{2P}\sum_{i=1}^{P}(r_i-1-\log r_i)$, where $r_i=F_i^{\mathrm{deg}}/F_i^{\mathrm{ref}}$. As the KL divergence is asymmetric, we define the final degradation measure as the symmetrized KL-divergence (\textbf{\texttt{SymKL}}),
$\mathcal{D}(x_{\mathrm{ref}},x_{\mathrm{deg}})=\frac12\mathrm{KL}(F^{\mathrm{ref}}\Vert F^{\mathrm{deg}})+\frac12\mathrm{KL}(F^{\mathrm{deg}}\Vert F^{\mathrm{ref}})$. The proposed measure satisfies $\mathcal{D}\ge0$, with equality if and only if the two Fisher profiles are identical. Larger values indicate greater reorganization of the parameter-sensitivity structure required for the frozen representation to explain the degraded signal. Because the formulation depends only on the coordinate-based representation and the reconstruction objective, the same measure applies unchanged across different signal modalities.

\vspace{-5pt}

\subsection{Exact closed-form Fisher for SIREN}
\label{sec:method:closed}

A direct evaluation of the empirical Fisher requires computing one parameter gradient for every coordinate, resulting in $N\times P$ gradient values for a signal with $N$ coordinates and an INR with $P$ parameters. For the signal resolutions considered in this work, for instance, corresponds to approximately $4\times10^{10}$ values for a single image, making the naive computation impractical due to its memory, and time requirements. However, we observed that, the diagonal Fisher admits an exact closed-form computation for coordinate MLPs. Let $a_n^{(\ell-1)}$ denote the input activation to layer $\ell$ at coordinate $n$, and let $\delta_n^{(\ell)}=\partial\ell_n/\partial z_n^{(\ell)}$ denote the corresponding backpropagated error. The gradient of the loss with respect to the weight matrix is then the outer product $\partial\ell_n/\partial W_\ell=\delta_n^{(\ell)}(a_n^{(\ell-1)})^\top$. Consequently, its elementwise square is separable, allowing the summation over all coordinates to factorize exactly as
$F_{W_\ell}=\frac{1}{N}\left(\Delta^{(\ell)}\odot\Delta^{(\ell)}\right)^\top\left(A^{(\ell-1)}\odot A^{(\ell-1)}\right)$,
where the rows of $A^{(\ell-1)}$ and $\Delta^{(\ell)}$ contain the activations $a_n^{(\ell-1)}$ and backpropagated errors $\delta_n^{(\ell)}$, respectively, for all coordinates. The diagonal Fisher for the bias parameters follows analogously as $F_{b_\ell}=\frac{1}{N}\sum_n\delta_n^{(\ell)}\odot\delta_n^{(\ell)}$.The required backpropagated errors are obtained using the standard backward recursion, with the sinusoidal activation contributing only through its derivative $\omega_\ell\cos(\omega_\ell z_n^{(\ell)})$. Consequently, the complete Fisher is computed using one forward pass, one backward pass, and one matrix multiplication per layer, without ever materializing per-coordinate parameter gradients. The complete derivation is provided in Supplementary materials.

\subsection{Physically grounded sensitivity protocol}
\label{sec:method:protocol}

Beyond assigning a degradation score, a useful measure should respond meaningfully as the severity of the underlying physical distortion increases. The rate of this response reflects the measure's ability to distinguish progressively degraded signals. We therefore introduce a physically grounded sensitivity measure that quantifies the relative change in a degradation measure with respect to the relative change in the underlying physical distortion. As existing measures differ in polarity, scale, and interpretation, their responses are not directly comparable. We therefore map every metric $m$ to a non-negative ratio-scale degradation value $\psi(m)$ with a true zero corresponding to identical signals. Similarity measures are transformed as $\psi(m)=1-m$, logarithmic fidelity measures such as PSNR, and SNR are converted to their corresponding linear error ratios using $\psi(m)=10^{-m/10}$, while measures that already represent non-negative degradation or distance values are left unchanged. Given a monotonic sequence of distortion severity levels
$d_1<\cdots<d_K$ (e.g., increasing blur standard deviation,
noise variance, erosion iterations, or reverberation time), and the corresponding transformed metric values $M_k=\psi(m_k)$, we define the step-wise sensitivity as $S_k=\left|\frac{(M_{k+1}-M_k)/M_k}{(d_{k+1}-d_k)/d_k}\right|$, which measures the percentage change in the degradation measure per percentage change in the underlying physical distortion. The overall sensitivity is obtained as the mean over all valid degradation steps, $S=\frac{1}{|\mathcal{V}|}\sum_{k\in\mathcal{V}}S_k$, where $\mathcal{V}$ denotes the set of valid consecutive steps. The same evaluation protocol is applied to every measure. Steps whose initial degradation value falls below a small noise threshold are excluded to avoid unstable ratios near zero, and distortion sequences that are non-monotonic in their physical parameters are evaluated as separate monotonic branches. This ensures that the reported sensitivity reflects the responsiveness of the degradation measure rather than artifacts of the evaluation protocol. Further details are provided in Supplementary materials.
\vspace{-7pt}
\section{Experiments}

We evaluate the proposed degradation measure on four representative signal modalities. RGB images, occupancy fields, audio, and medical imaging. These experiments demonstrate \textbf{\texttt{SymKL}}'s effectiveness in quantifying degradation across diverse domains while comparing against widely used modality-specific signal measures.

\subsection{Images}

\label{sec:exp:image}

We evaluate how faithfully the proposed measure tracks degradation severity on the KADID-10k \cite{lin2019kadid} dataset, which contains a diverse set of image degradations applied at multiple severity levels. For each reference image, an INR is fitted only to the pristine image using the method discussed above, and subsequently frozen. The reference and its degraded versions are then evaluated using the same learned representation, ensuring that the observed differences arise from degradation rather than additional optimization. We compare against commonly used full-reference image-quality measures, including PSNR, SSIM, MS-SSIM, LPIPS, CLIP-based distance, and DreamSim. As some of these measures represent similarity while others represent distance, for a fair evaluation their outputs are converted to a common degradation-oriented form before computing sensitivity as explained in sensitivity protocol section. 
Beyond commonly used image-domain measures, we investigate several weight-space measures that can be computed directly from the frozen reference INR without re-optimizing it on the degraded image. This controlled setting isolates the choice of weight-space representation from the effects of additional optimization. Let $F^{\mathrm{ref}}$ and $F^{\mathrm{deg}}$ denote the diagonal Fisher profiles induced by the reference and degraded images, and let $\bar{g}^{\mathrm{ref}}$ and $\bar{g}^{\mathrm{deg}}$ denote their corresponding mean-gradient vectors. We consider five alternatives. \textbf{Fisher cosine (F-Cos)} computes the cosine distance between $F^{\mathrm{ref}}$ and $F^{\mathrm{deg}}$, capturing changes in the relative distribution of Fisher values while remaining invariant to uniform scaling of their overall magnitudes. \textbf{Fisher relative $\ell_2$ (F-$\ell_2$)} computes the relative Euclidean difference, $\lVert F^{\mathrm{deg}}-F^{\mathrm{ref}}\rVert_2/\lVert F^{\mathrm{ref}}\rVert_2$, making it sensitive to changes in both the relative distribution and overall magnitude of the Fisher values. \textbf{Gradient cosine (G-Cos)} applies cosine distance to the mean-gradient vectors, measuring differences in the directions of the first-order responses. \textbf{Gradient relative $\ell_2$ (G-$\ell_2$)} computes the relative Euclidean difference between $\bar{g}^{\mathrm{ref}}$ and $\bar{g}^{\mathrm{deg}}$, quantifying changes in the magnitude of the first-order response. Finally, \textbf{Adam-step cosine (Adam-Cos)} computes the cosine distance between two first-step Adam update vectors, one induced by the reference image and the other by the degraded image, both evaluated from the same frozen reference parameters with the optimizer state initialized to zero.

\begin{table*}[!ht]
\centering
\caption{Mean relative sensitivity on KADID-10k distortions image-domain and weight-space baselines.
Best results are shown in bold and second-best results are underlined.
Higher is better.}
\label{tab:kadid_common_wins}
\scriptsize
\setlength{\tabcolsep}{3.0pt}
\renewcommand{\arraystretch}{1.15}

\resizebox{\textwidth}{!}{%
\begin{tabular}{
l
cccccc
@{\hspace{5pt}}
ccccc
>{\columncolor{green!12}}c
}
\toprule
&
\multicolumn{6}{c}{\textbf{Image-domain baselines}}
&
\multicolumn{6}{c}{\textbf{Weight-space baselines}}
\\

\cmidrule(lr){2-7}
\cmidrule(lr){8-13}

\textbf{Distortion}
& \textbf{PSNR}
& \textbf{SSIM}
& \textbf{MS-SSIM}
& \textbf{LPIPS}
& \textbf{CLIP}
& \textbf{DSim}
& \textbf{F-Cos}
& \textbf{F-$\ell_2$}
& \textbf{G-Cos}
& \textbf{G-$\ell_2$}
& \textbf{Adam-Cos}
& \cellcolor{green!18}\textbf{\texttt{SymKL}}
\\
\midrule

Motion blur
& 1.678 & 1.659 & 2.678 & 1.187 & 2.399 & \underline{2.860}
& 0.391 & 2.749 & 0.345 & 1.137 & 0.132
& \textbf{3.479}
\\


JPEG2000
& 1.418 & 0.843 & 1.154 & 1.363 & 1.506 & 1.558
& \underline{1.963} & 1.962 & 0.649 & 0.790 & 0.179
& \textbf{2.921}
\\


White noise
& 0.963 & 0.364 & 0.634 & 0.552 & 0.339 & 0.557
& 0.504 & \underline{0.974} & 0.396 & 0.512 & 0.120
& \textbf{1.169}
\\


Impulse noise
& \underline{1.004} & 0.799 & 0.827 & 0.515 & 0.334 & 0.702
& 0.461 & 1.004 & 0.452 & 0.580 & 0.118
& \textbf{1.345}
\\



Mean shift
& 3.059 & 2.382 & 3.222 & 3.172 & 2.787 & \underline{3.338}
& 0.402 & 2.990 & 0.052 & 1.068 & 0.034
& \textbf{3.462}
\\

Quantization
& \underline{2.340} & 1.244 & 1.753 & 1.721 & 1.309 & 1.610
& 1.127 & 2.236 & 0.112 & 0.901 & 0.069
& \textbf{2.862}
\\

Contrast
& 3.468 & 3.046 & \underline{3.815} & 3.227 & 3.318 & 3.663
& 0.413 & 3.804 & 0.076 & 1.195 & 0.048
& \textbf{4.359}
\\

\bottomrule
\end{tabular}%
}
\end{table*}

Table~\ref{tab:kadid_common_wins} compares the relative sensitivity of the proposed measure against both image-domain and weight-space baselines on representative KADID-10k distortion types. As can be seen across the distortions, \textbf{\texttt{SymKL}} consistently exhibits the highest relative sensitivity, indicating a stronger and more consistent response to degradation severity than existing alternatives. These results suggest that modeling degradation through Fisher-profile distributions captures distortion-induced changes more effectively than conventional image-domain metrics or other weight-space statistics extracted from the same frozen INR.

\begin{figure*}[!t]
    \centering
    \includegraphics[width=0.9\textwidth]{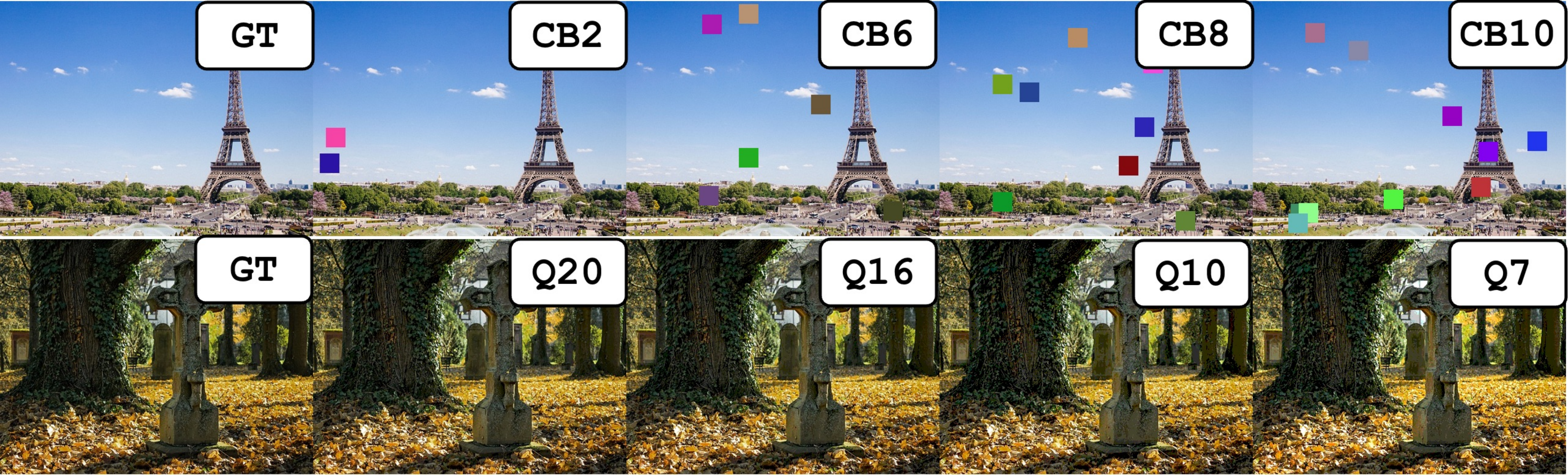}
\caption{Responses of different image-quality metrics to progressively increasing degradation severity. \textbf{Top:} increasing color blocks (2--10). \textbf{Bottom:} decreasing quantization levels (Q20--Q7). Existing metrics exhibit inconsistent or non-monotonic responses, whereas \textbf{\texttt{SymKL}} increases smoothly and monotonically with degradation severity.}
    \label{fig:progressive_response}
\end{figure*}

To further examine whether quality measures respond consistently to increasing degradation severity, we analyze two progressive distortion sequences from the KADID dataset corresponding to color-block corruption and intensity quantization (Fig.~\ref{fig:progressive_response}). The associated metric responses are summarized in Table~\ref{tab:progressive_response}. As each successive image represents a strictly stronger degradation than its predecessor, an ideal degradation measure should exhibit a monotonic response. However, several widely used image-quality metrics violate this expected behavior. For example, PSNR, SSIM, MS-SSIM, LPIPS, CLIP, and DreamSim all exhibit non-monotonic responses for at least one of the two degradation sequences, as indicated by the underlined entries in Table~\ref{tab:progressive_response}. In contrast, the proposed Fisher-based measure increases monotonically across both distortion sequences, producing a consistent response to progressively stronger degradations. These results suggest that the proposed representation provides a more stable and predictable characterization of degradation severity than both conventional image-domain measures and alternative learned perceptual metrics. Additionally, on a single NVIDIA RTX A5000 GPU, \textbf{\texttt{SymKL}} exhibits comparable evaluation time to DreamSim, requiring $52.36 \pm 0.05$ ms and $45.17 \pm 0.74$ ms per evaluation, respectively across three runs.

\begin{figure*}[!t]
    \centering
    \includegraphics[width=0.85\textwidth]{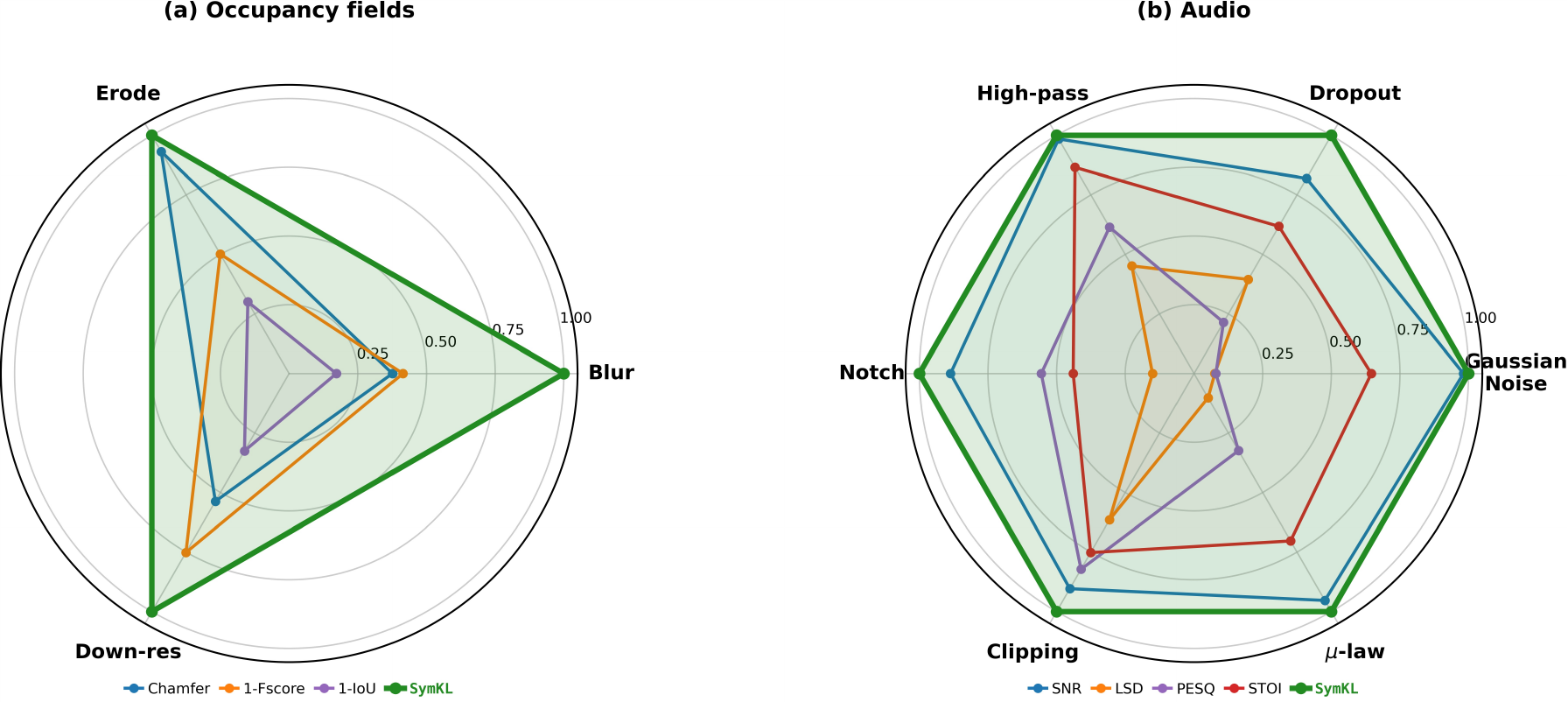}
    \caption{\textbf{Normalized sensitivity profiles across degradation types.} Each axis is normalized by the highest sensitivity attained for the corresponding degradation. Across both occupancy-field (left) and audio (right) degradations, \textbf{\texttt{SymKL}} consistently exhibits the strongest relative sensitivity, demonstrating a more responsive characterization of degradation than conventional modality-specific measures.}
    \label{fig:radar_profiles}
\end{figure*}


\begin{table}[!t]
\centering
\caption{Responses of different image-quality metrics to progressively increasing color-block degradation (CB) and progressively decreasing quantization levels (Q), both corresponding to increasing degradation severity. Underlined values indicate points at which a metric violates the expected monotonic response.}
\label{tab:progressive_response}
\scriptsize
\setlength{\tabcolsep}{2.15pt}
\renewcommand{\arraystretch}{1.1}

\begin{tabular}{lcccc@{\hspace{3pt}}cccc}
\toprule
&
\multicolumn{4}{c}{\textbf{Color Blocks (CB)}}
&
\multicolumn{4}{c}{\textbf{Quantization (Q)}} \\
\cmidrule(lr){2-5}
\cmidrule(lr){6-9}

\textbf{Metric}
& \texttt{\textbf{CB2}}
& \texttt{\textbf{CB6}}
& \texttt{\textbf{CB8}}
& \texttt{\textbf{CB10}}
& \texttt{\textbf{Q20}}
& \texttt{\textbf{Q16}}
& \texttt{\textbf{Q10}}
& \texttt{\textbf{Q7}} \\
\midrule

PSNR
& 25.865
& 23.131
& 21.295
& \underline{21.711}
& 31.529
& 29.515
& 25.421
& 23.884 \\

SSIM
& 0.989
& 0.970
& 0.964
& 0.958
& 0.956
& 0.932
& 0.856
& \underline{0.864} \\

MS-SSIM
& 0.981
& 0.920
& 0.874
& \underline{0.917}
& 0.990
& 0.983
& 0.957
& \underline{0.959} \\

LPIPS
& 0.045
& 0.135
& 0.196
& \underline{0.166}
& 0.017
& 0.029
& 0.078
& \underline{0.060} \\

CLIP
& 1.000
& 0.820
& \underline{0.835}
& 0.800
& 0.976
& \underline{0.982}
& 0.974
& 0.973 \\

DreamSim
& 0.129
& 0.243
& 0.331
& \underline{0.319}
& 0.007
& 0.011
& 0.025
& \underline{0.020} \\

\rowcolor{green!12}
\textbf{\texttt{SymKL}}
& \textbf{0.948}
& \textbf{1.957}
& \textbf{3.066}
& \textbf{3.126}
& \textbf{0.539}
& \textbf{0.911}
& \textbf{2.512}
& \textbf{4.227} \\

\bottomrule
\end{tabular}
\end{table}

\subsection{Occupancy Fields}

To evaluate whether the proposed degradation measure generalizes beyond natural images, we next consider occupancy fields representing 3D objects. We first generate the reference occupancy fields by following the construction procedure described in WIRE~\cite{saragadam2023wire}. We then progressively degrade these reference occupancy fields using three commonly encountered degradation types, Gaussian blur, morphological erosion, and resolution reduction. Blur is applied using Gaussian kernels with increasing standard deviations ($\sigma=\{1.0,1.5,2.0,3.0,4.0\}$), erosion using $1$--$5$ morphological iterations, and resolution reduction using downsampling factors of $\{1.5\times,2\times,3\times,4\times,6\times\}$. For each degradation sequence, we evaluate the response of the proposed measure and compare it against the geometric metrics commonly used for occupancy-field evaluation, namely Chamfer distance, F-score, and volumetric IoU. Left of \Cref{fig:radar_profiles} illustrates the max-normalized mean relative sensitivity over all degradation levels for the \emph{XYZ} occupancy field. As can be seen, across all three degradation types, the proposed measure consistently obtains the highest sensitivity, outperforming the conventional occupancy-field metrics by a clear margin. This demonstrates that the proposed approach captures progressive structural degradation more effectively (See Supplementary materials for the exact values).

\begin{figure*}[!t]
    \centering

    \begin{minipage}[c]{0.6\textwidth}
        \centering
        \includegraphics[width=\linewidth]{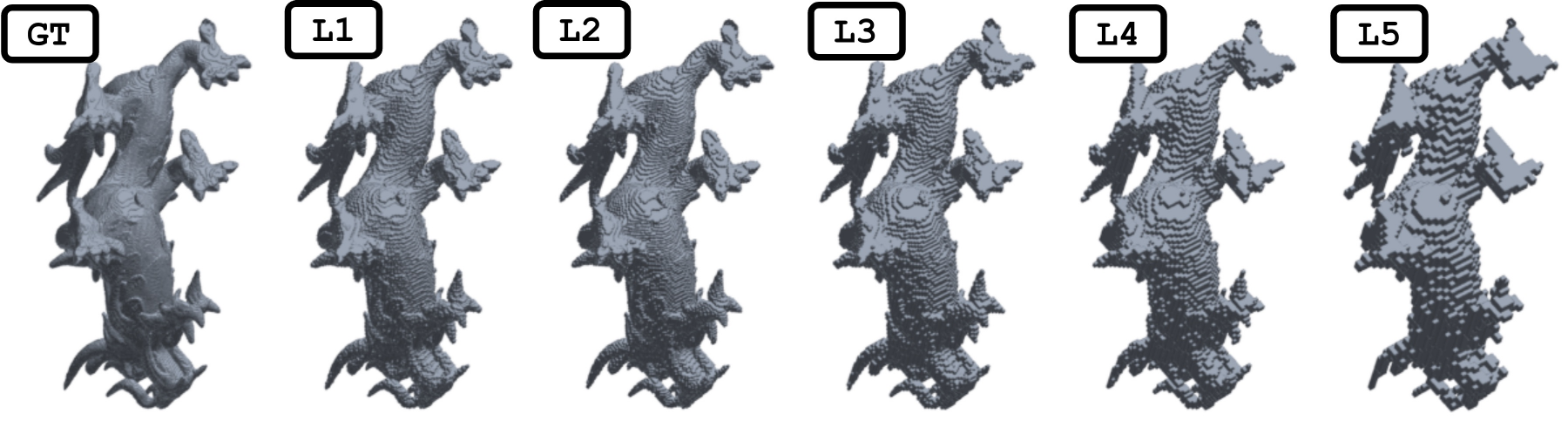}
        \captionof{figure}{
        Progressive downsampling of a representative occupancy field.
        From \texttt{GT} to \texttt{L5} (DS6), increasing downsampling
        progressively removes fine geometric details while preserving
        the overall structure.
        }
        \label{fig:occ_progressive}
    \end{minipage}
    \hfill
    \begin{minipage}[c]{0.385\textwidth}
        \centering
        \captionof{table}{
        Responses of occupancy-field quality measures under increasing
        downsampling.
        }
        \label{tab:occupancy_progression}

        \vspace{3pt}
        \scriptsize
        \setlength{\tabcolsep}{2.8pt}
        \renewcommand{\arraystretch}{1.45}

        \resizebox{\linewidth}{!}{%
        \begin{tabular}{lccccc}
            \toprule
            \textbf{Metric}
            & \textbf{DS1.5}
            & \textbf{DS2}
            & \textbf{DS3}
            & \textbf{DS4}
            & \textbf{DS6} \\
            \midrule
            Chamfer ($\times 10^{-5}$)
            & 2.910 & 4.072 & 5.118 & 9.062 & 17.860 \\

            F-score
            & 0.959 & 0.932 & 0.893 & 0.759 & 0.581 \\

            IoU
            & 0.960 & 0.945 & 0.933 & 0.905 & 0.866 \\

            \rowcolor{green!10}
            \textbf{\texttt{SymKL}}
            & \textbf{0.180}
            & \textbf{0.367}
            & \textbf{0.505}
            & \textbf{1.287}
            & \textbf{2.953} \\
            \bottomrule
        \end{tabular}%
        }
    \end{minipage}

    \vspace{-4pt}
\end{figure*}

To further investigate this behavior, \Cref{fig:occ_progressive} illustrates the progressive structural degradation of a representative occupancy field under increasing resolution reduction. The corresponding responses of the different quality measures are reported in \Cref{tab:occupancy_progression}. Unlike the image distortions considered in the previous subsection, all metrics preserve the correct ordering of the degradation levels, i.e., larger degradations consistently produce larger measured differences. However, correctly ordering degradations alone is insufficient if the response becomes increasingly compressed as degradation severity grows. In particular, volumetric IoU, one of the most widely adopted evaluation metrics for occupancy fields and INRs, exhibits a progressively attenuated response, with successive increases in degradation producing increasingly smaller relative changes in the metric. Similar behavior is observed for Chamfer distance and F-score, indicating that these measures increasingly underestimate the severity of additional structural degradation. A quantitative analysis of this attenuation is provided in the supplementary material. In contrast, \textbf{\texttt{SymKL}} maintains a strong and consistent response throughout the degradation sequence, preserving not only the correct ordering of progressively degraded occupancy fields but also clear separation between successive degradation levels. This experiment therefore reveals a complementary failure mode to that observed for natural images. Whereas image-quality metrics may fail by violating monotonicity under progressively stronger distortions, conventional occupancy-field metrics remain monotonic but exhibit an attenuated response that progressively underestimates additional degradation. The proposed approach addresses both behaviors by providing a stable, monotonic, and highly responsive characterization of degradation severity across occupancy fields.

\subsection{Audio}

To further evaluate the proposed degradation measure across signal modalities, we next consider audio signals. We use the \textit{dev-clean} split of the LibriSpeech dataset~\cite{panayotov2015librispeech}, from which we extract 54 reference clips of 3 seconds each (16\,kHz) spanning 33 speakers. Each reference clip is progressively degraded using six common audio degradation types, additive Gaussian noise with SNR levels of $\{30,25,20,15,10\}$\,dB, sample dropout with dropout rates of $\{2,5,10,20,35\}\%$, high-pass filtering with normalized cutoff frequencies of $\{0.02,0.05,0.10,0.20,0.40\}$, spectral notching with normalized notch widths of $\{0.02,0.05,0.10,0.20,0.35\}$, clipping with peak-amplitude thresholds of $\{0.80,0.50,0.30,0.15,0.07\}$, and $\mu$-law companding with quantization levels of $\{256,128,64,32,16\}$. This results in a total of 1,620 degraded audio samples. We compare the proposed measure against widely used full-reference audio quality metrics, including SNR, LSD, PESQ, and STOI.

The right figure of \Cref{fig:radar_profiles} illustrates the normalized-mean relative sensitivity across all degradation levels for each degradation type. As can be seen \textbf{\texttt{SymKL}} obtains the highest sensitivity across all six degradations, demonstrating a stronger and more consistent response to changes in audio quality than the conventional audio-specific metrics. The largest gains are observed for spectral notching, clipping, and $\mu$-law companding, while the proposed measure also attains the highest sensitivity for additive Gaussian noise, sample dropout, and high-pass filtering. These results indicate that \textbf{\texttt{SymKL}} provides a robust and modality-agnostic characterization of degradation that extends effectively from images and occupancy fields to audio signals.

\section*{Additional Results}

The supplementary material provides additional quantitative analyses supporting the main paper, including all referenced quantitative comparisons, extended qualitative visualizations, and experiments on medical imaging datasets, offering a broader evaluation across diverse signal modalities.

\section{Conclusion}
We introduced \textbf{\texttt{SymKL}}, a modality-agnostic representation-space degradation measure that quantifies signal degradation through the Fisher information geometry of a frozen implicit neural representation (INR). Unlike conventional full-reference measures, which rely on modality-specific representations, \textbf{\texttt{SymKL}} compares reference and degraded signals through a common Fisher representation derived from a shared INR. Because the measure depends only on a frozen reference representation and the reconstruction objective, rather than the underlying corruption process or its severity, it applies unchanged across modalities and naturally extends to signals for which no established full-reference measure exists. To enable efficient computation, we derived an exact closed-form expression for the diagonal Fisher of SIREN networks. Across images, audio, and 3D occupancy fields, \textbf{SymKL} consistently tracks degradation severity, remaining monotonic and discriminative even where conventional measures saturate or violate monotonicity. More broadly, this work suggests that learned neural representations themselves can serve as the substrate for signal analysis. Just as pixels, waveforms, and point clouds have traditionally formed the basis for degradation measures, future signal-processing algorithms may instead operate directly on the information geometry of learned representations, providing unified tools that naturally extend across diverse signal modalities.

\bibliography{aaai2027}

\clearpage
\newpage

\appendix

\section{Related Works}
\label{sec:related}

\paragraph{Implicit neural representations.}
INRs model signals as continuous functions that map spatial or temporal coordinates to signal values. Their ability to represent complex, high-frequency content is enabled through techniques such as positional encodings or specialized activation functions, including periodic nonlinearities~\cite{sitzmann2020implicit} and space-frequency compact nonlinearities~\cite{ramasinghe2022beyond, saragadam2023wire, jayasundara2025mire}. Owing to their flexible coordinate-based formulation, INRs have been successfully applied to a wide range of signal modalities, including neural radiance fields, occupancy fields, signed-distance fields, and neural compression, where the signal is represented by the parameters of a neural network~\cite{essakine2024we, jayasundara2026implicit}. Existing work primarily employs INRs as a representation for accurately reconstructing, compressing signals \cite{jayasundara2025sinr} or using them in downstream tasks \cite{jayasundara2026implicit, xu2022signal, jayasundara2026spectra}. In contrast, we use the learned INR as a representation to be analyzed rather than reconstructed, treating it as a common substrate on which a unified degradation measure can be defined across diverse signal modalities.

\paragraph{Learning in the weight space of neural representations.}
Recent work has increasingly explored learning directly from the parameters of neural representations rather than their outputs. Functa represents each fitted neural field by a learned modulation vector, which serves as a compact representation for downstream tasks~\cite{dupont2022data}, while inr2vec learns embeddings of INRs for shape understanding and retrieval~\cite{de2023deep}. More generally, weight-space learning methods process neural network parameters directly to learn representations of trained models, enabling tasks such as classification, retrieval, and self-supervised representation learning~\cite{qiu2026ensuring, yang2025weight, erkocc2023hyperdiffusion}. In contrast, our objective is not to learn representations from network parameters. Instead, we use the Fisher profile of a single reference INR to characterize how degradation alters the functional organization of a learned neural representation, yielding a degradation measure rather than a learned embedding.

\paragraph{Fisher information as a descriptor of neural representations.}
The Fisher information matrix characterizes the sensitivity of a model's predictions to perturbations of its parameters and has become a fundamental tool in statistical learning and optimization. It forms the basis of natural-gradient optimization~\cite{amari1998natural}, underpins Laplace approximations for Bayesian neural networks~\cite{ritter2018scalable}, and its diagonal has been widely adopted as a measure of parameter importance, most notably in elastic weight consolidation for continual learning~\cite{kirkpatrick2017overcoming}. Beyond optimization and uncertainty estimation, Fisher information has also been used as a representation of learned models. Closest to our work, Task2Vec represents learning tasks using the diagonal Fisher information of a fixed probe network after adaptation to each task, enabling task similarity to be quantified through comparisons of the resulting Fisher vectors~\cite{achille2019task2vec}. Our work builds on the same principle that parameter sensitivity reflects the information encoded by a neural network, but applies it in a fundamentally different setting. Rather than comparing datasets or learning tasks, we characterize degradation within a single signal using a fixed INR fitted to the pristine reference. The network parameters remain unchanged, and the Fisher profiles differ only through the target signal used for evaluation. Furthermore, we derive an exact closed-form expression for the diagonal Fisher information of SIREN, eliminating the expensive per-sample gradient computations required by conventional Fisher estimation.

\paragraph{Efficient INR fitting.}
Training an INR independently for every signal can be computationally expensive, motivating approaches that accelerate INR optimization. One direction employs hypernetworks to predict the parameters of a target network from an input representation~\cite{chauhan2024brief}. This idea has been extended to INRs using transformer-based weight decoders that directly predict INR parameters~\cite{chen2022transformers}. A complementary direction is meta-learning, which seeks parameter initializations that can be rapidly adapted to new signals with only a few optimization steps~\cite{finn2017model}. This mechanism has proven effective for INRs through learned initializations~\cite{tancik2021learned} and modulation-based adaptation~\cite{dupont2022coin++}. Our approach combines these two paradigms by employing a hypernetwork conditioned on DINOv2 features~\cite{oquab2024dinov2} to predict the weights of a SIREN, while meta-training the hypernetwork using a first-order MAML objective so that its predictions provide high-quality initializations for efficient test-time refinement.

\end{document}